\pdfoutput=1

\documentclass[10pt,twocolumn,letterpaper]{article}

\usepackage[pagenumbers]{cvpr} 

\usepackage{graphicx}
\usepackage{amsmath}
\usepackage{amssymb}
\usepackage{booktabs}

\usepackage{cellspace}
\usepackage{amsfonts}
\usepackage{physics} 
\usepackage[protrusion=true,tracking=true,kerning=true,expansion,final]{microtype}      
\usepackage{booktabs} 
\usepackage{subcaption}
\usepackage{nicefrac} 
\usepackage{cancel}

\usepackage{xspace}
\usepackage{upref}
\usepackage{textcomp}
\usepackage[T1]{fontenc}

\usepackage{array}
\usepackage{amssymb}
\usepackage{gensymb}
\usepackage{bbding}
\usepackage{mathtools}
\usepackage{centernot} 
\usepackage{bigints}
\usepackage{dsfont} 
\usepackage{esint} 
\usepackage{multirow}
\usepackage{enumitem}
\usepackage{graphbox} 
\usepackage{xcolor}

\usepackage{algorithm}
\usepackage{algpseudocode}
\usepackage{algorithmicx}
\usepackage[square,sort,comma,numbers]{natbib}

\usepackage{varioref} 

\usepackage[pagebackref=true,breaklinks=true,colorlinks,bookmarks=true]{hyperref}

\usepackage[capitalize]{cleveref}
\crefname{section}{Sec.}{Secs.}
\Crefname{section}{Section}{Sections}
\Crefname{table}{Table}{Tables}
\crefname{table}{Tab.}{Tabs.}

\crefname{appsec}{appendix}{appendices}
\Crefname{appsec}{Appendix}{Appendices}
\usepackage{url}

\renewcommand{\cite}[1]{\citep{#1}}
\definecolor{mydarkblue}{rgb}{0,0.08,0.45}
\definecolor{urlcolor}{rgb}{0,.145,.698}
\definecolor{linkcolor}{rgb}{.71,0.21,0.01}
\hypersetup{ %
	pdftitle={End-to-End Referring Video Object Segmentation with Multimodal Transformers},
	pdfauthor={Adam Botach, Evgenii Zheltonozhskii, Chaim Baskin},
	pdfsubject={},  
	pdfkeywords={},
	pdfborder=0 0 0,     
    pdfpagemode=UseOutlines,
	breaklinks=true,  
	colorlinks=true,
	bookmarksnumbered=true,
	urlcolor=urlcolor,
	linkcolor=linkcolor,
	citecolor=mydarkblue,
	filecolor=mydarkblue,
	pdfview=FitH}

\renewcommand*{\backref}[1]{} 
\renewcommand*{\backrefalt}[4]{%
	\ifcase #1 %
	\or
	(cited on p. #2)%
	\else
	(cited on pp. #2)%
	\fi
}
\makeatletter
\renewcommand{\@biblabel}[1]{#1.}
\makeatother

\vbadness=\maxdimen
\usepackage{bm}

\newcommand{\methodname}{MTTR}

\DeclareMathOperator*{\argmin}{arg\,min}  

\def\cvprPaperID{4222} 
\def\confName{CVPR}
\def\confYear{2022}

\begin{document}

\title{End-to-End Referring Video Object Segmentation with Multimodal Transformers}

\author{Adam Botach, Evgenii Zheltonozhskii, Chaim Baskin\\
    {Technion -- Israel Institute of Technology}\\
{\tt\small \{\href{mailto:botach@campus.technion.ac.il}{botach}, \href{mailto:evgeniizh@campus.technion.ac.il}{evgeniizh}\}@campus.technion.ac.il \quad 
\href{mailto:chaimbaskin@cs.technion.ac.il}{chaimbaskin@cs.technion.ac.il}}
}

\maketitle

\begin{abstract}
The referring video object segmentation task (RVOS) involves  segmentation of a text-referred object instance in the frames of a given video. Due to the complex nature of this multimodal task, which combines text reasoning, video understanding, instance segmentation and tracking, existing approaches typically rely on sophisticated pipelines in order to tackle it. In this paper, we propose a simple Transformer-based approach to RVOS. Our framework, termed Multimodal Tracking Transformer (\methodname{}), models the RVOS task as a  sequence prediction problem. Following recent advancements in computer vision and natural language processing, \methodname{} is based on the realization that video and text can be processed together effectively and elegantly by a single multimodal Transformer model. \methodname{} is end-to-end trainable, free of text-related inductive bias components and requires no additional mask-refinement post-processing steps. As such, it simplifies the RVOS pipeline considerably compared to existing methods. Evaluation on standard benchmarks reveals that \methodname{} significantly outperforms previous art across multiple metrics. In particular, \methodname{} shows impressive +5.7 and +5.0 mAP gains on the A2D-Sentences and JHMDB-Sentences datasets respectively, while processing 76 frames per second. In addition, we report strong results on the public validation set of Refer-YouTube-VOS, a more challenging RVOS dataset that has yet to receive the attention of  researchers. The code to reproduce our experiments is available at \url{https://github.com/mttr2021/MTTR}.
\end{abstract}


\section{Introduction}
\label{sec:intro}

\begin{figure}
\centering
\includegraphics[width=1.0\linewidth]{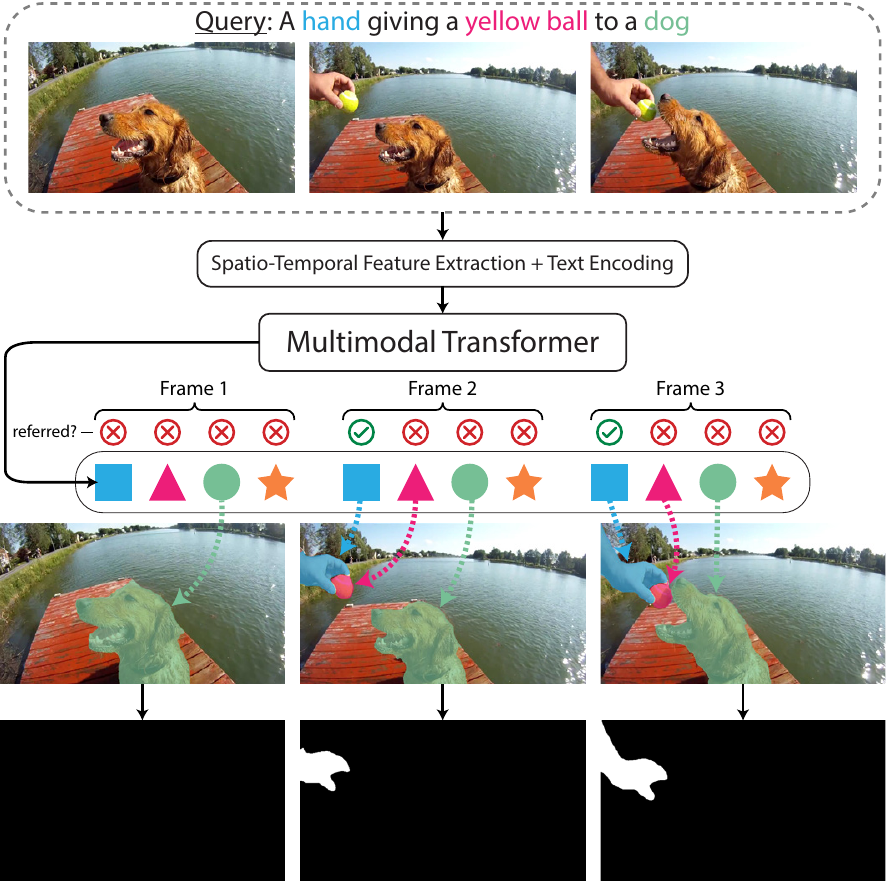}
 \caption{Given a text query and a sequence of video frames, the proposed model outputs prediction sequences for all object instances in the video prior to determining the referred instance. Here predictions with the same color and shape belong to the same sequence and attend to the same object instance in different frames. Note that the order of instance predictions for different frames remains the same. Best viewed in color.
 }
\vspace{-0.1cm}
\label{fig:task_visual}
\vspace{-0.55cm}
\end{figure}

Attention-based \cite{vaswani2017attention} deep neural networks exhibit impressive performance on various tasks across different fields, from computer vision \cite{ViT_2021,liu2021swin} to natural language processing \cite{Devlin2019BERTPO,brown2020gpt3}. These advancements make networks of this sort, such as the Transformer \cite{vaswani2017attention}, particularly interesting candidates for solving multimodal problems. By relying on the self-attention mechanism, which allows each token in a sequence to globally aggregate information from every other token, Transformers excel at modeling global dependencies and have become the cornerstone in most NLP tasks \cite{Devlin2019BERTPO, yang2019xlnet, radford2019language,brown2020gpt3}. Transformers have also started showing promise in solving computer vision tasks, from recognition \cite{ViT_2021} to object detection \cite{carion2020detr} and even outperforming the long-used CNNs as general-purpose vision backbones \cite{liu2021swin}.

The referring video object segmentation task (RVOS) involves the segmentation of a text-referred object instance in the frames of a given video. Compared with the 
referring image segmentation task (RIS) \cite{yu2016refcoco,mao2016refcocoplus}, in which objects are mainly referred to by their appearance, in RVOS objects can also be referred to by the actions they are performing or in which they are involved. This renders RVOS significantly harder 
than RIS, as text expressions that refer to actions often cannot be properly deduced from a single static frame. Furthermore, unlike their image-based counterparts, RVOS methods may be required to establish data association of the referred object across multiple frames (tracking) in order to deal with disturbances such as occlusions or motion blur. 

To solve these challenges and effectively align video with text, existing RVOS approaches \cite{hui2021cstm,liu2021cmpc,ning2020polar} typically rely on complicated pipelines. In contrast, here we propose a simple, end-to-end Transformer-based approach to RVOS. Using recent advancements in Transformers for textual feature extraction \cite{vaswani2017attention,liu2019roberta}, visual feature extraction \cite{ViT_2021,liu2021swin, liu2021vswin} and object detection \cite{carion2020detr,wang2021vistr}, we develop a framework that significantly outperforms existing approaches.
To accomplish this, we employ a single multimodal Transformer and model the task as a sequence prediction problem. Given a video and a text query, our model generates prediction sequences for \textit{all} objects in the video before determining the one the text refers to. Additionally, our method is free of text-related inductive bias modules and utilizes a simple cross-entropy loss to align the video and the text. As such, it is much less complicated than previous approaches to the task.

The proposed pipeline is schematically depicted in \cref{fig:task_visual}. 
First, we extract linguistic features from the text query using a standard Transformer-based text encoder, and visual features from the video frames using a spatio-temporal encoder. The features are then passed into a multimodal Transformer, which outputs several sequences of object predictions \cite{wang2021vistr}. Next, to determine which of the predicted sequences best corresponds to the referred object, we compute a text-reference score for each sequence. For this we propose a \textit{temporal segment voting scheme} that allows our model to focus on more relevant parts of the video when making the decision.

Our main contributions are as follows:
\begin{itemize}
\item We present a Transformer-based RVOS framework, dubbed \textbf{M}ultimodal \textbf{T}racking \textbf{Tr}ansformer (\methodname{}), which models the task as a parallel sequence prediction problem and outputs predictions 
for \textit{all} objects in the video prior to selecting the one referred to by the text.
\item Our sequence selection strategy is based on a \textit{temporal segment voting scheme}, a novel reasoning scheme that allows our model to focus on more relevant parts of the video with regards to the text.
\item The proposed method is end-to-end trainable, free of text-related inductive bias modules, and requires no additional mask refinement. 
As such, it greatly simplifies the RVOS pipeline compared to existing approaches.
\item We thoroughly evaluate our method. 
On the A2D-Sentences and JHMDB-Sentences \cite{gavrilyuk2018a2d}, \methodname{} significantly outperforms all existing methods across all metrics. We also show strong results on the public validation set of Refer-YouTube-VOS \cite{seo2020urvos}, a challenging dataset that has yet to receive attention in the literature.
\end{itemize}

\section{Related Work}
\label{sec:related}
\paragraph{Referring video object segmentation.}
The RVOS task was introduced by \citet{gavrilyuk2018a2d}, whose goal was to attain pixel-level segmentation of actors and their actions in video content. 
To effectively aggregate and align visual, temporal and lingual information from video and text, state-of-the-art RVOS approaches typically rely on complicated pipelines  \cite{wang2019acga,wang2020cmd,ning2020polar,McIntosh_2020_CVPR,liu2021cmpc}. \citet{gavrilyuk2018a2d} proposed an I3D-based \cite{carreira2017i3d} encoder-decoder architecture that generated dynamic filters from text features and convolved them with visual features to obtain the 
masks. Following them, \citet{wang2020cmd} added spatial context to the kernels with deformable convolutions \cite{Dai_2017_ICCV}. 
For a more effective representation, 
VT-Capsule \cite{McIntosh_2020_CVPR} encoded each modality in capsules \cite{capsule_networks_nips_2017}, while ACGA \cite{wang2019acga} utilized a co-attention mechanism to enhance the multimodal features.
To improve positional relation representations in the text, PRPE \cite{ning2020polar} explored a positional encoding mechanism based on polar coordinates. URVOS \cite{seo2020urvos} improved tracking capabilities by performing language-based object segmentation on a key frame 
and then propagating its 
mask throughout the video. 
AAMN \cite{yang2020aamn} utilized a top-down approach where an off-the-shelf object detector is used to localize objects in the video prior to parsing relations between visual and textual features.  CMPC-V \cite{liu2021cmpc} achieved state-of-the-art results by constructing a 
temporal graph from video and text features, and applying graph convolution \cite{graph_conf_ICLR_2017} to detect the referred entity.


\begin{figure*}
\centering
\includegraphics[width=1.0\linewidth]{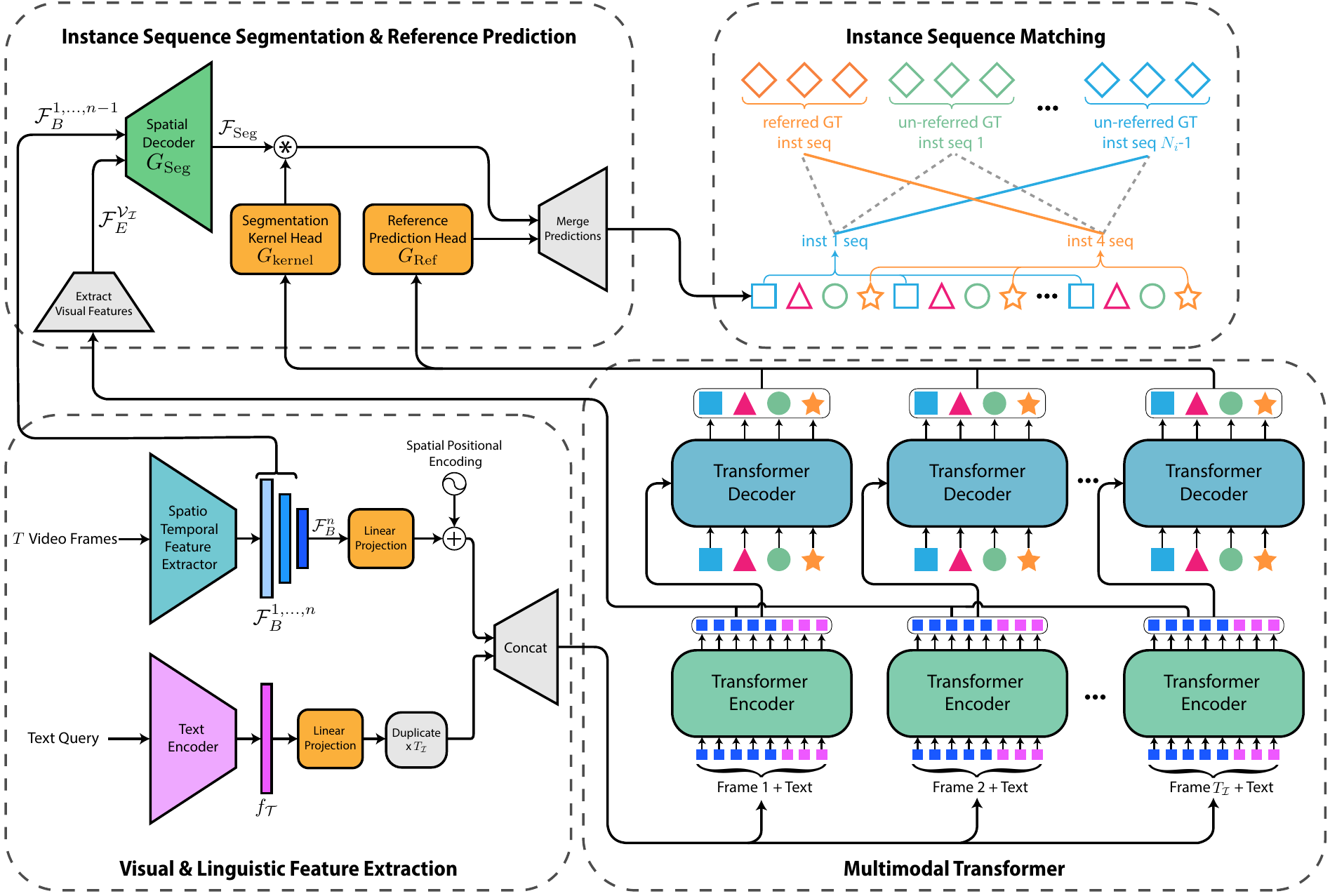}
 \caption{A detailed overview of \methodname{}. First, the input text and video frames are passed through feature encoders and then concatenated into multimodal sequences (one per frame). A multimodal Transformer then encodes the feature relations and decodes instance-level features into a set of prediction sequences. Next, corresponding mask and reference prediction sequences are generated. Finally, the predicted sequences are matched with the ground truth sequences for supervision (in training) or used to generate the final prediction (during inference).}
\label{fig:approach}
\vspace{-0.5cm}
\end{figure*}

\paragraph{Transformers.} 
The Transformer \cite{vaswani2017attention} was introduced as an attention-based building block for sequence-to-sequence machine translation, and since then has become the cornerstone for most NLP tasks \cite{Devlin2019BERTPO, yang2019xlnet,radford2019language,brown2020gpt3}. Unlike previous architectures, the Transformer relies entirely on the attention mechanism to draw dependencies between input and output. 

Recently, the introduction of Transformers 
to computer vision tasks has demonstrated spectacular performance. DETR \cite{carion2020detr}, which utilizes a non-auto-regressive 
Transformer, 
simplifies the traditional object detection pipeline while achieving performance comparable to that of CNN-based detectors \cite{faster_rcnn_NIPS_2017}. Given a fixed set of learned object queries, DETR reasons about the global context of an image and the relations between its objects and then 
outputs a final set of detection predictions in parallel.
VisTR \cite{wang2021vistr} extends the idea behind DETR to video instance segmentation. It views the task as a direct end-to-end parallel sequence prediction problem. By supervising video instances at the sequence level as a whole, VisTR is able to output an ordered sequence of masks for each instance in a video  directly (i.e., natural tracking).

ViT \cite{ViT_2021} introduced the Transformer to image recognition by using linearly projected 
patches as tokens for a Transformer encoder. Swin Transformer \cite{liu2021swin} proposed a general-purpose backbone for computer vision based on a hierarchical Transformer whose representations are computed inside shifted windows. This architecture was also 
extended to the video domain \cite{liu2021vswin}, which we adapt as our temporal encoder. 

Another recent relevant work is MDETR \cite{kamath2021mdetr}, a DETR-based end-to-end multimodal detector that detects objects in an image conditioned on a text query. Different from our method, their approach is designed to work 
on static images, and its performance largely depends on well-annotated datasets that contain \textit{aligned} text and box annotations, the types of which are not available in the RVOS task.

\section{Method}
\label{sec:method}



\subsection{Method Overview}
\label{sec:overview}

\paragraph{Task definition.} The input of RVOS consists of a frame sequence $\mathcal{V} = \{v_i\}_{i=1}^{T}$, where $v_i \in \mathbb{R}^{C\times H_{0}\times W_{0}}$, and a text query $\mathcal{T}=\{t_i\}_{i=1}^L$, where $t_i$ is the $i^\text{th}$ word in the text. Then, for a subset of frames of interest $\mathcal{V_{\mathcal{I}}}\subseteq\mathcal{V}$ of size $T_\mathcal{I}$, the goal is to segment the object referred by $\mathcal{T}$ in each frame in $\mathcal{V_I}$. 
We note that since producing mask annotations requires significant  efforts, $\mathcal{V_{\mathcal{I}}}$ rarely contains all of the frames in $\mathcal{V}$.

\paragraph{Feature extraction.} 
We begin by extracting features from each frame in the sequence $\mathcal{V}$ using a deep spatio-temporal encoder. Simultaneously, linguistic features are extracted from the text query $\mathcal{T}$ using a Transformer-based \cite{vaswani2017attention} text encoder. Then, the spatio-temporal and linguistic features are linearly projected to a shared dimension $D$.

\paragraph{Instance prediction.}
In the next step, the features of each frame of interest are flattened and separately concatenated with the text embeddings, producing a set of $T_\mathcal{I}$ multimodal sequences. These sequences are fed in parallel into a Transformer \cite{vaswani2017attention,carion2020detr}. In the Transformer's encoder layers, the textual embeddings and the visual features of each frame exchange information. Then, the decoder layers, which are fed with $N_{\text{q}}$ object queries per input frame, query the multimodal sequences for entity-related information and store it in the object queries. Corresponding queries of different frames share the same trainable weights and are trained to attend to the same instance in the video (each one in its designated frame). We refer to these queries (represented by the same unique color and shape in \cref{fig:task_visual,fig:approach}) as queries belonging to the same \textit{instance sequence}. This design allows for natural tracking of each object instance in the video \cite{wang2021vistr}.

\paragraph{Output generation.}
For each output instance sequence,  
we generate a a corresponding 
mask sequence using an FPN-like \cite{lin2017fpn} spatial decoder and dynamically generated conditional convolution kernels \cite{tian2020conditional,wang2021maxdeeplab}. Finally, we use a novel text-reference score function that, based on text associations, determines which of the object query sequences has the strongest association with the object described in $\mathcal{T}$, and returns its 
segmentation sequence as the model's prediction.

\subsection{Temporal Encoder} \label{subsec:backbone}
A suitable temporal encoder for the RVOS task should be able to extract \textit{both} visual characteristics (e.g., shape, size, location) and action semantics for each instance in the video. Several previous works \cite{gavrilyuk2018a2d,ning2020polar,liu2021cmpc} utilized the Kinetics-400 \cite{kay2017kinetics} pre-trained I3D network \cite{carreira2017i3d} as their temporal encoder. However, since I3D was originally designed for action classification, using its outputs as-is for tasks that require fine details (e.g., instance segmentation) is not ideal as the features it outputs tend to suffer from spatial misalignment caused by temporal downsampling. To compensate for this side effect, past state-of-the-art approaches came up with different solutions, from auxiliary mask refinement algorithms \cite{krahenbuhl2012crf, liu2021cmpc} to utilizing additional backbones that operate alongside the temporal encoder \cite{hui2021cstm}. In contrast, our end-to-end approach does not require any additional mask refinement steps and utilizes a single backbone.

Recently, the Video Swin Transformer \cite{liu2021vswin} was proposed as a generalization of the Swin Tranformer \cite{liu2021swin} to the video domain. While the original Swin was designed with dense predictions (such as segmentation) in mind, Video Swin was tested mainly on action recognition benchmarks. To the best of our knowledge, we are the first to utilize it (with a slight modification) for video segmentation.
As opposed to I3D, Video Swin contains just a single temporal downsampling layer and can be easily modified to output \textit{per-frame} feature maps (we refer to \cref{supp_temporal_encoder} for more details). As such, it is a much better choice for processing a full sequence of consecutive video frames for segmentation purposes.

\subsection{Multimodal Transformer} \label{subsec:Transformer}
For each frame of interest, the temporal encoder generates a feature map $f_{t}^\mathcal{V_\mathcal{I}}\in \mathbb{R}^{H \times W \times C_{\mathcal{V}}}$ and the text encoder outputs a linguistic embedding vector $f_{\mathcal{T}}\in \mathbb{R}^{L \times D_{\mathcal{T}}}$ for the text. These visual and linguistic features are linearly projected to a shared dimension $D$. The features of each frame are then flattened and \textit{separately} concatenated 
with the text embeddings, resulting in a set of $T_\mathcal{I}$ multimodal sequences, each of shape $(H \times W + L) \times D$. 
The multimodal sequences along with a set of $N_q$ instance sequences are then fed in parallel into a Transformer as described earlier. Our Transformer architecture is similar to the one used in DETR \cite{carion2020detr}. Accordingly, the problem now comes down to finding the instance sequence that attends to the text-referred object.

\subsection{The Instance Segmentation Process} \label{subsec:segmentation}
Our segmentation process, as shown in \cref{fig:approach}, consists of several steps.
First, given $\mathcal{F}_{E}$, the updated multimodal sequences output by the last Transformer encoder layer, we extract and reshape the video-related part of each sequence (i.e., the first $H\times W $ tokens)  into the set $\mathcal{F}_E^{\mathcal{V}_\mathcal{I}}$.
Then, we take $\mathcal{F}_B^{1,\dots, n-1}$, the outputs of the first $n-1$ blocks of our temporal encoder, and hierarchically fuse them with $\mathcal{F}_E^{\mathcal{V}_\mathcal{I}}$ using an FPN-like \cite{lin2017fpn} spatial decoder $G_\text{Seg}$. This process results in semantically-rich, high resolution feature maps of the video frames, denoted as $\mathcal{F}_\text{Seg}$.
\begin{align}
\mathcal{F}_\text{Seg}=
\qty{f^t_{\text{Seg}}}_{t=1}^{T_\mathcal{I}}, 
\: f^t_{\text{Seg}}\in \mathbb{R}^{D_{s}\times \frac{H_0}{4} \times \frac{W_0}{4}} \label{eq:seg}
\end{align}
Next, for each instance sequence $\mathcal{Q}=\{q_t\}_{t=1}^{T_{\mathcal{I}}}, q_t \in \mathbb{R}^{D}$ output by the Transformer decoder, we use a two-layer perceptron $G_\text{kernel}$ to generate a corresponding sequence of conditional segmentation kernels \cite{tian2020conditional,wang2021maxdeeplab}.
\begin{align}
{G_\text{kernel}(\mathcal{Q}}) = \{k_t\}_{t=1}^{T_{\mathcal{I}}}, k_t \in \mathbb{R}^{D_s}
\end{align}
Finally, a sequence of segmentation masks $\mathcal{M}$ is generated for $\mathcal{Q}$ by convolving each segmentation kernel with its corresponding frame features, followed by a bilinear upsampling operation to resize the masks into ground-truth resolution,
\begin{align}
\resizebox{.9\hsize}{!}{$\mathcal{M}=\{m_t\}_{t=1}^{T_{\mathcal{I}}}, m_t = \text{Upsample}(k_t * f^t_{\text{Seg}})\in \mathbb{R}^{{H_0} \times W_0}. $}
\end{align}
\subsection{Instance Sequence Matching} \label{subsec:matching}
During the training process we need to determine which of the predicted instance sequences best fits the referred object. 
However, if the video sequence $\mathcal{V}$ contains additional annotated instances, we found that supervising their detection (as negative examples) alongside that of the referred instance 
helps stabilize the training process. 

Let us denote by $y$ the set of ground-truth sequences that are available for $\mathcal{V}$, and by $\hat{y}=\{\hat{y}_i\}_{i=1}^{N_{q}}$ the set of the predicted instance sequences. We assume that the number of predicted sequences ($N_q$) is chosen to be strictly greater than the number of annotated instances (denoted $N_i$) and that the ground-truth sequences set is padded with $\varnothing$ (no object) to fill any missing slots. Then, we want to find a matching between the two sets \cite{carion2020detr,wang2021vistr}. Accordingly, we search for a permutation $\hat{\sigma} \in S_{N_{q}}$ with the lowest total cost:
\begin{align}
\hat{\sigma} = \argmin_{\sigma \in S_{N_\text{q}}} \sum_{i=1}^{N_q}\mathcal{C}_{\text{Match}}\qty(\hat{y}_{\sigma(i)},y_i),
\end{align}
where $\mathcal{C}_\text{Match}$ is a pair-wise matching cost.
The optimal permutation $\hat{\sigma}$ can be  computed efficiently using the Hungarian algorithm \cite{kuhn1955hungarian}. 
Each ground-truth sequence is of the form
\begin{align}
y_i =(m_i, r_i)= \qty(\{m_i^t\}_{t=1}^{T_\mathcal{I}}, \{r_i^t\}_{t=1}^{T_\mathcal{I}}), \label{eq:pred}
\end{align}
where $m_i^t$ is a ground-truth mask, and $r_i^t\in \{0, 1\}^2$ is a one-hot referring vector, i.e., the positive class means that $y_i$ corresponds to the text-referred object \textit{and} that this object is visible in the corresponding video frame $v_t$. Note that if $y_i$ is a padding sequence then $m_i = \varnothing$.

To allow our model to produce reference predictions in the form of \cref{eq:pred}, we use a reference prediction head, denoted $G_\text{Ref}$, which consists of a single linear layer of shape $D\times 2$ followed by a softmax layer. Given a predicted object query $q\in \mathbb{R}^D$, this head takes $q$ as input and outputs a reference prediction $\hat{r} \equiv G_\text{Ref}(q)$.

Thus, each prediction of our model is a pair of sequences:
\begin{align}
\hat{y}_{j} =\qty(\hat{m}_{j}, \hat{r}_{j})= \qty(\{\hat{m}_{j}^t\}_{t=1}^{T_\mathcal{I}}, \{\hat{r}_{j}^t\}_{t=1}^{T_\mathcal{I}}).
\end{align}
We define the pair-wise matching cost function as the sum 
\begin{align}
\resizebox{0.905\hsize}{!}{$
\mathcal{C}_{\text{Match}}(\hat{y}_{j},y_i) = \mathds{1}_{\{m_i \ne \varnothing\}} \Big[ \lambda_d\mathcal{C}_{\text{Dice}}(\hat{m}_{j},m_i) + \lambda_r\mathcal{C}_\text{Ref}(\hat{r}_{j},r_i) \Big], \label{eq:cost}
$}
\end{align}
where $\lambda_d, \lambda_r \in \mathbb{R}$ are hyperparameters.
$\mathcal{C}_{\text{Dice}}$ supervises the predicted mask sequence using the ground-truth mask sequence by averaging the negation of the Dice coefficients \cite{milletari2016vnet} of each pair of corresponding masks at every time step. We refer to \cref{loss_and_cost_appendix} for the full definition of this cost function.
$\mathcal{C}_\text{Ref}$ supervises the 
reference predictions using the corresponding ground-truth sequence as follows
\begin{align}
\mathcal{C}_\text{Ref}(\hat{r}_{j},r_i) =-\frac{1}{T_\mathcal{I}}\sum_{t=1}^{T_\mathcal{I}} \hat{r}_j^t \cdot r_{i}^t.
\end{align}

\begin{table*}
	\centering
    \setlength\extrarowheight{0.5pt}
\begin{tabular}{@{\extracolsep{4pt}}l ccccc cc c@{}} 
\toprule	 	
	\multirow{2}{*}{\textbf{Method}} &\multicolumn{5}{c}{\textbf{Precision}} &\multicolumn{2}{c}{\textbf{IoU}} & \multirow{2}{*}{\textbf{mAP}}  \\ \cline{2-6}\cline{7-8}
    & \textbf{50\%} & \textbf{60\%} & \textbf{70\%} & \textbf{80\%} & \textbf{90\%}   & \textbf{Overall}   & \textbf{Mean}  & \\
\midrule	
\citet{hu2016segmentation} &34.8 & 23.6& 13.3 &3.3 & 0.1&47.4 &35.0 & 13.2\\
\citet{gavrilyuk2018a2d} (RGB) &47.5 & 34.7 & 21.1 & 8.0 & 0.2 &53.6 & 42.1 & 19.8 \\
RefVOS \cite{bellver2020refvos}  &57.8& -- & -- & -- & 9.3 &67.2 &49.7 & --\\
AAMN \cite{yang2020aamn} &68.1 & 62.9 &52.3& 29.6& 2.9&61.7& 55.2& 39.6 \\
CMSA+CFSA \cite{ye2021cfsa} &48.7& 43.1 &35.8 &23.1 &5.2&61.8& 43.2& --\\
CSTM \cite{hui2021cstm}  & 65.4& 58.9 &49.7 &33.3 &9.1 & 66.2 & 56.1 & 39.9\\
CMPC-V (I3D) \cite{liu2021cmpc} &65.5 & 59.2 & 50.6 & 34.2 & 9.8 &65.3& 57.3& 40.4\\
\midrule
\methodname{} ($w=8$, ours) &72.1&68.4&60.7&45.6&16.4&70.2&61.8&44.7\\
\methodname{} ($w=10$, ours) &\textbf{75.4}&\textbf{71.2}&\textbf{63.8}&\textbf{48.5}&\textbf{16.9}&\textbf{72.0}&\textbf{64.0}&\textbf{46.1}\\
		\bottomrule
	\end{tabular}
	\vspace{-0.2cm}
	\caption
		{Comparison with state-of-the-art methods on A2D-Sentences \cite{gavrilyuk2018a2d}.
		}
	\label{tbl:a2ds}
	\vspace{-0.3cm}
\end{table*}

\begin{table*}
	\centering
    \setlength\extrarowheight{0.5pt}
\begin{tabular}{@{\extracolsep{4pt}}l ccccc c cc@{}} 
\toprule	 	
	\multirow{2}{*}{\textbf{Method}} &\multicolumn{5}{c}{\textbf{Precision}} &\multicolumn{2}{c}{\textbf{IoU}} & \multirow{2}{*}{\textbf{mAP}}  \\ \cline{2-6}\cline{7-8}
    & \textbf{50\%} & \textbf{60\%} & \textbf{70\%} & \textbf{80\%} & \textbf{90\%}   & \textbf{Overall}   & \textbf{Mean}  & \\
\midrule	
\citet{hu2016segmentation} &63.3 &35.0 &8.5 &0.2 &0.0 &54.6& 52.8& 17.8\\
\citet{gavrilyuk2018a2d} (RGB)  &69.9 & 46.0 & 17.3& 1.4 &0.0&54.1 &54.2 & 23.3 \\
AAMN \cite{yang2020aamn}  &77.3 & 62.7& 36.0& 4.4 & 0.0 &58.3 & 57.6 & 32.1 \\
CMSA+CFSA \cite{ye2021cfsa}  &76.4 &62.5 &38.9 &9.0&\textbf{0.1}&62.8 &58.1 & --\\
CSTM \cite{hui2021cstm}  &78.3 & 63.9 & 37.8 & 7.6 & 0.0 &59.8 & 60.4 & 33.5  \\
CMPC-V (I3D)  \cite{liu2021cmpc} &81.3 & 65.7& 37.1 &7.0 &0.0&61.6 &61.7& 34.2\\
\midrule
\methodname{} ($w=8$, ours) &91.0&81.5&57.0&14.4&\textbf{0.1}&67.4&67.9&36.6\\
\methodname{} ($w=10$, ours) &\textbf{93.9}&\textbf{85.2}&\textbf{61.6}&\textbf{16.6}&\textbf{0.1}&\textbf{70.1}&\textbf{69.8}&\textbf{39.2}\\
		\bottomrule
	\end{tabular}
	\vspace{-0.2cm}
	\caption{ Comparison with state-of-the-art methods on JHMDB-Sentences  \cite{gavrilyuk2018a2d}.
		}
	\label{tbl:jhmdbs}
	\vspace{-0.6cm}
\end{table*}

\subsection{Loss Functions} \label{subsec:loss_fns}
Let us denote (with a slight abuse of notation) by $\hat{y}$ the set of predicted instance sequences permuted according to the optimal permutation $\hat{\sigma} \in S_{N_{q}}$. Then, we can define our loss function as follows:
\begin{align}
\resizebox{.90\hsize}{!}{$
\mathcal{L}(\hat{y},y) =
\sum_{i=1}^{N_q} \mathds{1}_{\{m_i \ne \varnothing\}} \mathcal{L}_\text{Mask}(\hat{m}_{i},m_i) 
+\mathcal{L}_\text{Ref}(\hat{r}_{i},r_i).
$}
\end{align}
Following VisTR \cite{wang2021vistr}, the first term, dubbed $\mathcal{L}_\text{Mask}$, ensures mask alignment between the predicted and ground-truth sequences. As such, this term is defined as a combination of the Dice \cite{milletari2016vnet} and the per-pixel Focal \cite{lin2017focal} loss functions:
 \begin{align}
\resizebox{.88\hsize}{!}{$ \displaystyle \mathcal{L}_\text{Mask}(\hat{m}_{i},m_i) =
\lambda_d\mathcal{L}_\text{Dice}(\hat{m}_{i},m_i) + 
\lambda_f\mathcal{L}_\text{Focal}(\hat{m}_{i},m_i)  ,$
}
 \end{align}
where $\lambda_d, \lambda_f \in \mathbb{R}$ are hyperparameters. Both $\mathcal{L}_\text{Dice}$ and $\mathcal{L}_\text{Focal}$ are applied on corresponding masks at every time step, and are normalized by the number of instances inside the training batch. We refer to \cref{loss_and_cost_appendix} for the full definitions of these functions.

The second loss term, denoted $\mathcal{L}_\text{Ref}$, is a cross-entropy term that supervises the sequence reference predictions:
 \begin{align}
\mathcal{L}_\text{Ref}(\hat{r}_{i},r_i) &=  -\lambda_r \frac{1}{T_\mathcal{I}}\sum_{t=1}^{T_\mathcal{I}} r_i^t \cdot \log(\hat{r}_{i}^t),
 \end{align}
where $\lambda_r \in \mathbb{R}$ is a hyperparameter. In practice we further downweight the terms of the negative (``unreferred'') class by a factor of 10 to account for class imbalance \cite{carion2020detr}. Also, note that the same $\lambda_r$ and $\lambda_d$ are used as weights in the matching cost (\ref{eq:cost}) and loss functions.
Intriguingly, despite $\mathcal{L}_\text{Ref}$'s simplicity and lack of explicit text-related inductive bias, it was able to deliver equivalent or even better performance compared with more complex loss functions \cite{kamath2021mdetr} that we tested. Hence, and for the sake of simplicity, no additional loss functions are used for text supervision in our method.

 \subsection{Inference} \label{subsec:inference}
For a given sample of video and text, let us denote by $\mathcal{R}=\{\hat{r}_i\}_{i=1}^{N_q}$ the set of reference prediction sequences output by our model. Additionally, we denote by $p_\text{ref}(\hat{r}_i^t)$ the probability of the positive (``referred'') class for a given reference prediction $\hat{r}_i^t$. During inference we return the segmentation mask sequence $\mathcal{M}_\text{pred}$ that corresponds to $ \hat{r}_\text{pred}$, the predicted reference sequence with the highest positive score:
 \begin{align}
 \hat{r}_\text{pred} = \operatorname*{arg\,max}_{\hat{r}_i \in \mathcal{R}} \sum_{t=1}^{T_\mathcal{I}} p_\text{ref}(\hat{r}_i^t).
 \end{align}
This sequence selection scheme, which we term the \textit{``temporal segment voting scheme''} (TSVS), grades each prediction sequence based on the total association of its terms with the text referred object. Thus, it allows our model to focus on more relevant parts of the video (in which the referred object is visible), and disregard less relevant parts (which may depict irrelevant objects or in which the referred object is occluded) when making the decision. We refer to \cref{tsvs_effect_analysis} for further analysis of the effect of TSVS.

\section{Experiments}
\label{sec:experiment}

To evaluate our approach, we conduct experiments on three referring video object segmentation datasets.
The first two, A2D-Sentences and JHMDB-Sentences \cite{gavrilyuk2018a2d}, were created by adding textual annotations to the original A2D \cite{xu2015can} and JHMDB \cite{jhuang2013towards} datasets. Each video in A2D has 3--5 frames annotated with pixel-level segmentation masks, while in JHMDB, 2D articulated human puppet masks are available for all frames.
We refer to \cref{appendix_a2d_jhmdb} for more details.
We adopt Overall IoU, Mean IoU, and precision@K to evaluate our method on these datasets. \textit{Overall IoU} computes the ratio between the total intersection and the total union area over all the test samples. \textit{Mean IoU} is the averaged IoU over all the test samples. 
\textit{Precision@K} considers the percentage of test samples whose IoU scores are above a threshold K, where $K \in [0.5, 0.6, 0.7, 0.8, 0.9]$. We also compute mean average precision (mAP) over 0.50:0.05:0.95 \cite{lin2014coco}. 

We want to note that we found inconsistencies in the mAP metric calculation in previous studies. For example, examination of published code revealed incorrect calculation of the metric as the average of the precision@K metric over several K values.
To avoid further confusion and ensure a fair comparison, we suggest adopting the COCO API\footnote{\url{https://github.com/cocodataset/cocoapi}} for mAP calculation.  For reference, a full implementation of the evaluation that utilizes the API is released with our code.

We further evaluate \methodname{} on the more challenging Refer-YouTube-VOS dataset, introduced by \citet{seo2020urvos}, who provided textual annotations for the original YouTube-VOS dataset \cite{xu2018youtube}. Each video has pixel-level instance segmentation annotations for every fifth frame. The original release of Refer-YouTube-VOS contains two subsets. 
One subset contains \textit{first-frame expressions} that describe only the first frame. 
The other contains \textit{full-video expressions} that are based on the whole video and are, therefore, 
more challenging. Following the introduction of the RVOS competition\footnote{\url{https://youtube-vos.org/dataset/rvos/}}, only the more challenging subset of the dataset  is publicly available now. 
Since ground-truth annotations are available only for the training samples and the test server is currently inaccessible, we report results on the validation samples by uploading our predictions to the competition’s server\footnote{\url{https://competitions.codalab.org/competitions/29139}}.
We refer to \cref{appendix_refer_youtube_vos} for more details.
The primary evaluation metrics for this dataset are the average of the region similarity ($\mathcal{J}$) and the contour accuracy ($\mathcal{F}$) \cite{perazzi2016benchmark}.

\subsection{Implementation Details}
As our temporal encoder we use the smallest (``tiny'') 
Video Swin Transformer \cite{liu2021vswin} pretrained on Kinetics-400  \cite{kay2017kinetics}. The original Video Swin 
consists of four blocks with decreasing spatial resolution. We found the output of the fourth block to be too small for small object detection and hence we only utilize the first three blocks. 
We use the output of the third block as the input of the multimodal Transformer, while the outputs of the earlier blocks are fed into the spatial decoder. We also modify the encoder's single temporal downsampling layer to output \textit{per-frame} feature maps as required by our model. 
As our text encoder we use the Hugging Face \cite{wolf2020Transformers} implementation of RoBERTa-base \cite{liu2019roberta}. For A2D-Sentences \cite{gavrilyuk2018a2d} we feed the model windows of $w=8$ frames with the annotated target frame in the middle. 
Each frame is resized such that the shorter side is at least 320 pixels and the longer side is at most 576 pixels. 
For Refer-YouTube-VOS \cite{seo2020urvos}, we use windows of $w=12$ consecutive annotated frames during training, and full-length videos (up to 36 annotated frames) during evaluation.
Each frame is resized such that the shorter side is at least 360 pixels and the longer side is at most 640 pixels. 
We do not use any segmentation-related pretraining, e.g., on COCO \cite{lin2014coco}, which is known to boost segmentation performance \cite{wang2021vistr}.  We refer the reader to \cref{supp_additional_imp_det} for more implementation details.

\begin{table}
	\centering
    \setlength\extrarowheight{0.3pt}
\begin{tabular}{@{\extracolsep{4pt}}l ccc @{}} 
\toprule	 	
	\textbf{Method} & \textbf{$\mathcal{J} \& \mathcal{F} $} & \textbf{$\mathcal{J}$} & \textbf{$\mathcal{F}$}   \\ 
\midrule	
URVOS \cite{seo2020urvos} &47.23 &45.27 &49.19\\
CMPC-V (I3D) \cite{liu2021cmpc} & 47.48 &45.64 & 49.32 \\ \midrule
 \citet{ding2021progressive}$^+$  &	54.8 &53.7 &56.0 \\ 
\methodname{} (ours) & \textbf{55.32} & \textbf{54.00} & \textbf{56.64}\\
		\bottomrule
	\end{tabular}
	\caption
		{ Results on Refer-YouTube-VOS. The upper half is evaluated on the original validation set, while the bottom half is evaluated on the public validation set. $^+$ -- ensemble. 
		}
	\label{tbl:yt}
	\vspace{-0.35cm}
\end{table}

\begin{figure*}
\centering
\includegraphics[width=1.0\linewidth]{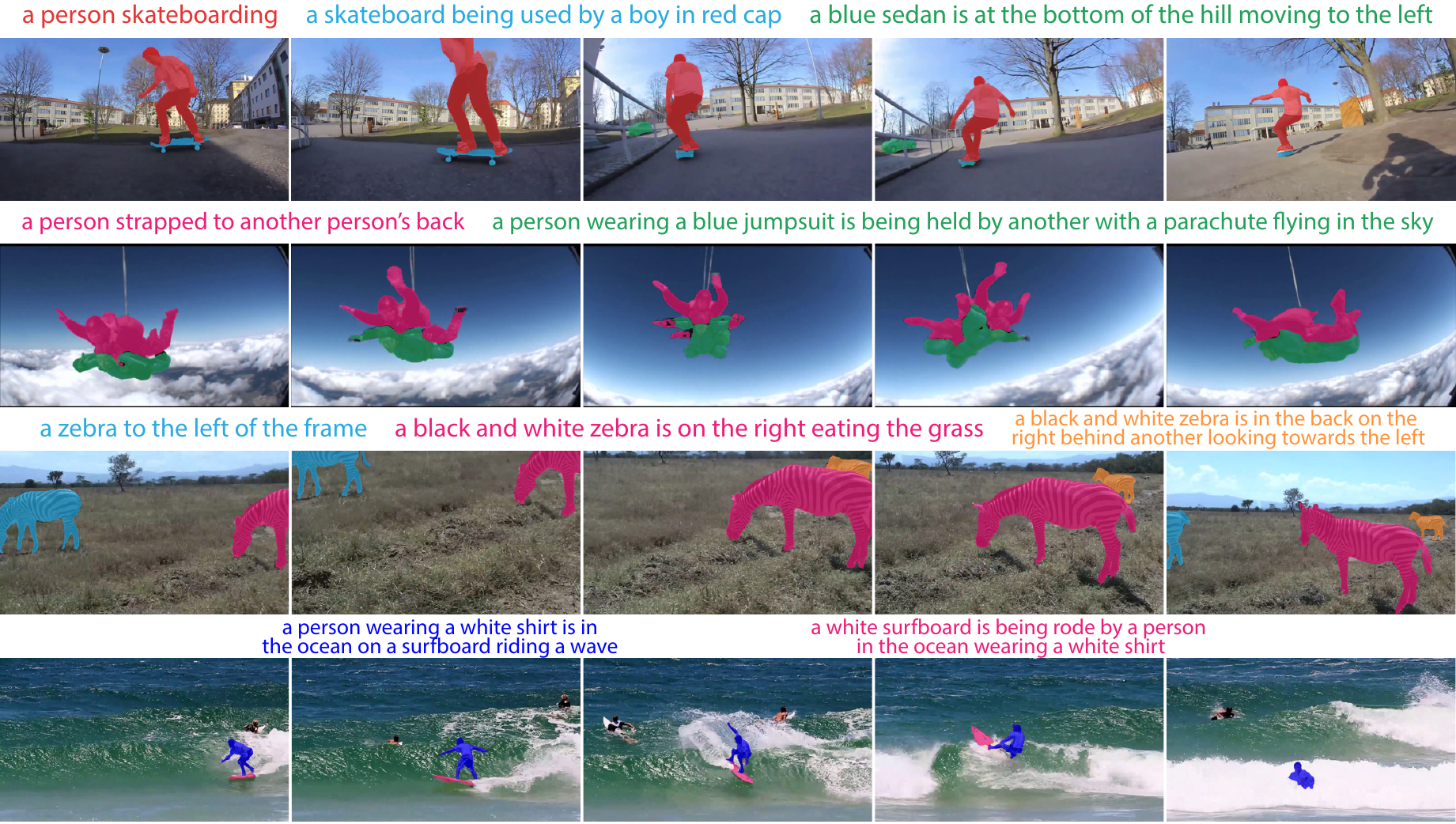}
	\vspace{-0.7cm}
 \caption{Visual examples of \methodname's performance on the Refer-YouTube-VOS \cite{seo2020urvos} validation set. Best viewed in color.}
\label{fig:example}
	\vspace{-0.35cm}
\end{figure*}

\begin{table*}
	\centering
    \setlength\extrarowheight{0.3pt}
  \subfloat[][Non-temporal backbones.]{
   \resizebox{0.35\linewidth}{!}{
   \begin{tabular}{@{\extracolsep{4pt}}l c cc @{}} 
\toprule	 	
	\multirow{2}{*}{\textbf{Method}}  &\multicolumn{2}{c}{\textbf{IoU}} & \multirow{2}{*}{\textbf{mAP}}   \\ \cline{2-3}
    &  \textbf{Overall}   & \textbf{Mean}  & \\
    \midrule
CMPC-I \cite{liu2021cmpc} &64.9&51.5&35.1\\
\methodname{} (DeepLab-ResNet101) &67.5&60.2&41.2\\
\methodname{} (Video Swin-T, $w=1$) &68.9&60.3&41.8\\
		\bottomrule
	\end{tabular}
	}
	\label{tbl:non-temporal}}%
	\hfill
	\subfloat[][Input window size effect.]{ 
   \resizebox{0.25\linewidth}{!}{
   \begin{tabular}{@{\extracolsep{4pt}}l c cc c@{}} 
\toprule	 	
	\multirow{2}{*}{$\mathbf{w}$}  &\multicolumn{2}{c}{\textbf{IoU}} & \multirow{2}{*}{\textbf{mAP}}    \\ \cline{2-3}
    &  \textbf{Overall}   & \textbf{Mean}   \\
\midrule	
$w=1$ &68.9&60.3&41.8\\
$w=4$ &69.7&61.5&43.8\\
$w=6$ &69.5&61.8&44.0\\
$w=8$ &70.2&61.8&44.7\\
$w=10$ &\textbf{72.0}&\textbf{64.0}&\textbf{46.1}\\
$w=12$ &69.3&62.0&44.0\\
		\bottomrule
	\end{tabular}
	}
	\label{tbl:ablation_window}
	}%
	\hfill
  \subfloat[][Word embeddings effect.]{
   \resizebox{0.35\linewidth}{!}{
   \begin{tabular}{@{\extracolsep{4pt}}l c cc@{}} 
\toprule	 	
	\multirow{2}{*}{\textbf{Method}}  &\multicolumn{2}{c}{\textbf{IoU}} & \multirow{2}{*}{\textbf{mAP}}  \\ \cline{2-3}
    &  \textbf{Overall}   & \textbf{Mean}  & \\
\midrule	
RoBERTa (base) &69.5&61.8&44.0\\
BERT (base) &69.7&62.1&44.3\\
Distill-RoBERTa (base) &70.5&62.4&43.8\\
GloVe &68.3&60.9&43.4\\
fastText &67.3&59.5&43.1\\
		\bottomrule
	\end{tabular}
	}
	\label{tbl:ablation_text}
	}
	\vspace{-0.1cm}
    \caption{Ablation studies on A2D-Sentences \cite{gavrilyuk2018a2d} dataset. }
	\label{tbl:ablation}
	\vspace{-0.5cm}
\end{table*}

\subsection{Comparison with State-of-the-Art Methods}
We compare our method with existing approaches on the A2D-Sentences dataset. 
For fair comparison with existing works \cite{hui2021cstm, liu2021cmpc}, our model is trained and evaluated for this purpose with windows of size 8.
As shown in \cref{tbl:a2ds}, our method significantly outperforms existing approaches across all metrics. 
For example, our model shows a 
4.3 mAP gain over current state of the art, and an absolute improvement of 6.6\% on the most stringent metric P@0.9, which demonstrates its ability to generate high-quality masks. We also note that our top configuration ($w=10$) achieves a massive 5.7 mAP gain and 6.7\% absolute improvement on both Mean and Overall IoU compared to the current state of the art. Impressively, this configuration is able to do so while processing 76 frames per second on a single RTX 3090 GPU.

Following previous works \cite{gavrilyuk2018a2d, liu2021cmpc}, we evaluate the generalization ability of our model 
by evaluating it on JHMDB-Sentences without fine-tuning. We uniformly sample three frames from each video and evaluate our best model on these frames. As shown in \cref{tbl:jhmdbs}, our method generalizes well and 
outperforms all existing approaches. 
Note that all methods (including ours) produce low results on P@0.9. This can be attributed to JHMDB's \cite{jhuang2013towards} imprecise mask annotations which were generated by a coarse human puppet model.

Finally, we report our results on the public validation set of Refer-YouTube-VOS \cite{seo2020urvos} 
in \cref{tbl:yt}. As mentioned earlier, this subset contains only the more challenging \textit{full-video expressions} from the original release of Refer-YouTube-VOS. 
Compared with existing methods \cite{seo2020urvos,liu2021cmpc} which trained and evaluated on the full version of the dataset, our model demonstrates superior performance across all metrics despite being trained on less data and evaluated exclusively on a more challenging subset. Additionally, our method shows competitive performance compared with the methods that led in the 2021 RVOS competition \cite{liang2021rethinking,ding2021progressive}. We note, however, that these methods use ensembles and are trained on additional segmentation and referring datasets  \cite{lin2014coco,xu2018youtube,yu2016refcoco,mao2016refcocoplus}.

\subsection{Ablation Studies}
We conduct 
ablation studies on A2D-Sentences to evaluate our model's design and robustness. 
Unless stated otherwise, we use window size $w=6$. 
An ablation study on the number of object queries can be found in \cref{appendix_ablations}.

\paragraph{Temporal encoder.} To evaluate \methodname's performance independently of the temporal encoder, we compare it with CMPC-I, the image-targeted version of CMPC-V \cite{liu2021cmpc}. Following CMPC-I, we use DeepLab-ResNet101 \cite{chen2017deeplab} pretrained on PASCAL-VOC \cite{everingham2010pascal} as a visual feature extractor. We train our model using only the target frames (i.e., without additional frames for temporal context). As shown in \cref{tbl:non-temporal}, our method significantly surpasses CMPC-I across all metrics, with a 6.1 gain in mAP and 8.7\% absolute improvement in Mean IoU. In fact, 
this configuration of our model surpasses all existing methods \textit{regardless} of the temporal context. 

\paragraph{Temporal context.} In \cref{tbl:ablation_window} we study the effect of the temporal context size on \methodname's performance. A larger temporal context enables better extraction of action-related information.
For this purpose, we train and evaluate our model using different window sizes. As expected, widening the temporal context leads to large performance gains, with an mAP gain of 4.3 and an absolute Mean IoU improvement of 3.7\% when gradually changing the window size from 1 to 10. Intriguingly, however, peak performance on A2D-Sentences is obtained using $w=10$, as widening the window even further (e.g., $w=12$) results in a performance drop.

\paragraph{Text encoder.} To study the effect of the selected word embeddings on our model's performance, we train our model using two additional widely-used Transformer-based text encoders, namely BERT-base \cite{Devlin2019BERTPO} and Distill-RoBERTa-base \cite{sanh2019distilbert}, a distilled version of RoBERTa \cite{liu2019roberta}. Additionally, we experiment with GloVe \cite{pennington2014glove} and fastText \cite{bojanowski2017enrichingfasttext}, two simpler word embedding methods. As shown in \cref{tbl:ablation_text}, our model achieves comparable performance when relying on the different Transformer-based encoders, which demonstrates its robustness to this change. Unsurprisingly, however, performance is slightly worse when relying on the simpler methods. This may be explained by the fact that while Transformer-based encoders are able to dynamically encode sentence context within their output embeddings, simpler methods disregard this context and merely rely on fixed pretrained embeddings.

\paragraph{Supervision of un-referred instances.} To study the effect of supervising the detections of un-referred instances alongside that of the referred instance in each sample, we train different configurations of our model without supervision of un-referred instances. Intriguingly, in all such experiments our model immediately converges to a local minimum of the text loss ($\mathcal{L}_\text{Ref}$), where the same object query is repeatedly matched with all ground-truth instances, thus leaving the rest of the object queries untrained. In some experiments our model manages to escape this local minimum after a few epochs and then achieves comparable performance with our original configuration. Nevertheless, in other experiments this phenomenon significantly hinders its final mAP score. 

\subsection{Qualitative Analysis}
As illustrated in \cref{fig:example}, \methodname{} can successfully track and segment the referred objects even in challenging situations where they are surrounded by similar instances, occluded, or completely outside of the frame in large parts of the video.
\section{Conclusion}
\label{sec:conclusion}
We introduced \methodname, a simple Transformer-based approach to RVOS that models the task as a sequence prediction problem. Our end-to-end method considerably simplifies existing RVOS pipelines by simultaneously processing both text and video frames in a single multimodal Transformer. Extensive evaluation of our approach on standard benchmarks reveals that our method outperforms existing state-of-the-art methods by a large margin (e.g., a 5.7 mAP improvement on A2D-Sentences). We hope our work will inspire others to see the potential of Transformers for solving complex multimodal tasks. 

\clearpage

{\small
\bibliographystyle{ieee_fullname}
\bibliography{refvos}
}

\clearpage
\appendix{}
\renewcommand\thefigure{\thesection.\arabic{figure}} 
\renewcommand\thetable{\thesection.\arabic{table}} 
\renewcommand\theequation{\thesection.\arabic{equation}}  
\setcounter{figure}{0}  
\setcounter{table}{0}

\crefalias{section}{appsec}
\crefalias{subsection}{appsec}
\crefalias{subsubsection}{appsec}

\section{Additional Method Details}
\subsection{Loss and Cost Functions}
\label{loss_and_cost_appendix}
In this section we present the full definitions of the Dice \cite{milletari2016vnet} and Focal \cite{lin2017focal} cost and loss functions used in \cref{subsec:matching} and \cref{subsec:loss_fns}.

Let $m_\text{GT}=\{m^t_\text{GT}\}_{t=1}^{T_\mathcal{I}}$ and $m_\text{Pred}=\{m^t_\text{Pred}\}_{t=1}^{T_\mathcal{I}}$ be ground-truth and prediction mask sequences respectively. Also, denote by $N_p$ the number of pixels a mask has and by $p_i \in \mathbb{R}$ the $i^\text{th}$ pixel in a given mask.

\subsubsection{Dice Cost and Loss}
Given two segmentation masks $m_A, m_B$, the Dice coefficient \cite{milletari2016vnet} between the two masks is defined as follows:
\begin{align}
\mathrm{DICE}(m_A, m_B) = \frac{2 * |m_A \cap m_B|}{|m_A| + |m_B|}, 
\end{align}
where for a mask $m$, $|m| = \sum_{i=1}^{N_p}{p_i}$ and $|m_A \cap m_B| = \sum_{i=1}^{N_p}{p^A_i \cdot p^B_i}$. Note that in practice we also add a smoothing constant $s = 1$ to both the numerator and denominator of the above expression to avoid possible division by $0$.

Given the above, the Dice cost $\mathcal{C}_\text{Dice}$ between the mask sequences $m_\text{GT}$ and $m_\text{Pred}$ is defined as
\begin{align}
    \resizebox{.83\linewidth}{!}{$
    \mathcal{C}_\text{Dice}(m_\text{GT}, m_\text{Pred}) = -\frac{1}{T_\mathcal{I}}\sum_{t=1}^{T_\mathcal{I}}{\mathrm{DICE}(m^t_\text{GT}, m^t_\text{Pred})}
    $}.
\end{align}

Similarly, the Dice loss $\mathcal{L}_\text{Dice}$ between the two sequences is defined as
\begin{equation}
    \resizebox{.83\linewidth}{!}{$
    \mathcal{L}_\text{Dice}(m_\text{GT}, m_\text{Pred}) = \sum_{t=1}^{T_\mathcal{I}}{1 - \mathrm{DICE}(m^t_\text{GT}, m^t_\text{Pred})}
    $}.
\end{equation}

\subsubsection{Focal Loss}
The Focal loss \cite{lin2017focal} between two corresponding segmentation masks $m_\text{GT}^t$ and $m_{Pred}^t$ for time step $t$ is defined as
\begin{align}
    \resizebox{.84\linewidth}{!}{$
    \mathrm{FL}(m_\text{GT}^t, m_{Pred}^t) = \frac{1}{N_p}\sum_{i=1}^{N_p}{-\alpha^{T}_i(1-p^{T}_i)^{\gamma}\log(p^{T}_i)},
    $}
\end{align}
where $p^{T}_i$ is the probability predicted for the ground-truth class of the $i^\text{th}$ pixel:
\begin{align}
    p^\text{T}_i = \begin{cases}
    p^\text{Pred}_i & p^\text{GT}_i = 1 \\
    1 - p^\text{Pred}_i & \text{otherwise},
    \end{cases}
\end{align}
and $\alpha^{T}_i \in [0, 1]$ is a class balancing factor defined as
 \begin{align}
    \alpha^{T}_i = \begin{cases}
    \alpha & p^\text{GT}_i = 1 \\
    1 - \alpha & \text{otherwise}.
    \end{cases}
\end{align}

Following \cite{lin2017focal, wang2021vistr} we use $\alpha=0.25, \gamma=2$. We refer to \cite{lin2017focal} for more information about these hyperparameters.

Given the above, the Focal loss $\mathcal{L}_\text{Focal}$ between the ground-truth and predicted mask sequences $m_\text{GT}$ and $m_\text{Pred}$ is defined as
\begin{align}
    \mathcal{L}_\text{Focal}(m_\text{GT}, m_\text{Pred}) = \sum_{t=1}^{T_\mathcal{I}}{\mathrm{FL}(m_\text{GT}^t, m_{Pred}^t)}.
\end{align}

\section{Additional Dataset Details}
\subsection{A2D-Sentences \& JHMDB-Sentences}
\label{appendix_a2d_jhmdb}
A2D-Sentences \cite{gavrilyuk2018a2d} contains 3,754 videos (3,017 train, 737 test) with 7 actors classes performing 8 action classes. Additionally, the dataset contains 6,655 sentences describing the actors in the videos and their actions. JHMDB-Sentences \cite{gavrilyuk2018a2d} contains 928 videos along with 928 corresponding sentences describing 21 different action classes. 

\subsection{Refer-YouTube-VOS}
\label{appendix_refer_youtube_vos}
The original release of Refer-YouTube-VOS \cite{seo2020urvos} contains 27,899 text expressions for 7,451 objects in 3,975 videos. The objects belong to 94 common categories. The subset with the \textit{first-frame expressions} contains 10,897 expressions for 3,412 videos in the train split and 1,993 expressions for 507 videos in the validation split. The subset with the \textit{full-video expressions} contains 12,913 expressions for 3,471 videos in the train split and 2,096 expressions for 507 videos in the validation split. Following the introduction of the RVOS competition\footnote{\url{https://youtube-vos.org/dataset/rvos/}}, only the more challenging \textit{full-video expressions} subset is publicly available now, so we use this subset exclusively in our experiments. Additionally, this subset's original validation set was split into two separate competition validation and test sets of 202 and 305 videos respectively. Since ground-truth annotations are available only for the training set and the test server is currently closed, we report results exclusively on the competition validation set by uploading our predictions to the competition’s server\footnote{\url{https://competitions.codalab.org/competitions/29139}}.

\section{Additional Implementation Details}
\label{supp_additional_imp_det}
\subsection{Temporal Encoder Modifications}
\label{supp_temporal_encoder}
The original architecture of Video Swin Transformer \cite{liu2021vswin} contains a single temporal down-sampling layer, realized as a 3D convolution with kernel and stride of size $2\times 4 \times 4$ (the first dimension is temporal). However, since our multimodal Transformer expects per-frame embeddings, we removed this temporal down-sampling step by modifying the kernel and stride of the above convolution to size $1\times 4 \times 4$. In order to achieve this while maintaining support for the Kinetics-400 \cite{kay2017kinetics} pretrained weights of the original Swin configuration, we summed the pretrained kernel weights of the aforementioned convolution on its temporal dim, resulting in a new $1\times 4 \times 4$ kernel. This solution is equivalent to (but more efficient than) duplicating each frame in the input sequence before inserting it into the temporal encoder.

\subsection{Multimodal Transformer}
We employ the same Transformer architecture proposed by \citet{carion2020detr}. The decoder layers are fed with a set of $N_q = 50$ object queries per input frame. For efficiency reasons we only utilize 3 layers in both the encoder and decoder, but note that more layers may lead to additional performance gains, as demonstrated by \citet{carion2020detr}. Also, similarly to \citet{carion2020detr}, fixed sine spatial positional encodings are added to the features of each frame before inserting them into the Transformer. No positional encodings are used for the text embeddings, as in our experiments using sine embeddings have led to reduced performance and learnable encodings had no effect compared to using no encodings at all.

\subsection{Instance Segmentation}
The spatial decoder $G_\text{Seg}$ is an FPN-like \cite{lin2017fpn} module consisting of several 2D convolution, GroupNorm \cite{wu2018group} and ReLU layers. Nearest neighbor interpolation is used for the upsampling steps. The segmentation kernels and the feature maps of $\mathcal{F}_\text{Seg}$ are of dimension $D_s = 8$ following \cite{tian2020conditional}. 

\subsection{Additional Training Details}
We use $D=256$ as the feature dimension of the multimodal Transformer's inputs and outputs. The hyperparameters for the loss and matching cost functions are $\lambda_r = 2, \lambda_d = 5, \lambda_f = 2$. 

Following \citet{carion2020detr} we utilize AdamW \cite{loshchilov2017decoupled} as the optimizer with weight decay set to $10^{-4}$ during training. We also apply gradient clipping with a maximal gradient norm of $0.1$. A learning rate of $10^{-4}$ is used for the Transformer and $5\cdot 10^{-5}$ for the temporal encoder. The text encoder is kept frozen.

Similarly to \citet{carion2020detr} we found that utilizing auxiliary decoding losses on the outputs of all layers in the Transformer decoder expedites training and improves the overall performance of the model.

During training, to enhance model's position awareness, we randomly flip the input frames horizontally and swap direction-related words in the corresponding text expressions accordingly (e.g., the word 'left' is replaced with 'right').

We train the model for 70 epochs on A2D-Sentences \cite{gavrilyuk2018a2d}. The learning rate is decreased by a factor of 2.5 after the first 50 epochs. In the default configuration we use window size $w=8$ and batch size of 6 on 3 RTX 3090 24GB GPUs. Training takes about 31 hours in this configuration.
On Refer-YouTube-VOS \cite{seo2020urvos} the model is trained for 30 epochs, and the learning rate is decreased by a factor of 2.5 after the first 20 epochs. In the default configuration we use window size $w=12$ and batch size of 4 on 4 A6000 48GB GPUs. Training takes about 45 hours in this configuration. 

\section{Additional Experiments}

\subsection{Ablations}
\label{appendix_ablations}
\paragraph{Number of object queries.}
To study the effect of the number of object queries on \methodname's performance, we train and evaluate our model on A2D-Sentences using window size $w = 6$ and different values of $N_q$. As shown in \cref{tbl:ablation_num_queries}, the best performance is achieved for $N_q=50$. Our hypothesis is that when using lower values of $N_q$ the resulting set of object queries may not be diverse enough to cover a large set of possible object detections. On the other hand, using higher values of $N_q$ may require a longer training schedule to obtain good results, as the probability of each query being matched with a ground-truth instance (and thus updated) at each training iteration is lower. 

\begin{table}
	\centering
    \setlength\extrarowheight{0.3pt}
   \begin{tabular}{@{\extracolsep{4pt}}l c cc c@{}} 
\toprule	 	
	\multirow{2}{*}{$\bm{N_q}$}  &\multicolumn{2}{c}{\textbf{IoU}} & \multirow{2}{*}{\textbf{mAP}}    \\ \cline{2-3}
    &  \textbf{Overall}   & \textbf{Mean}   \\
\midrule	
$10$ &\textbf{70.1}&61.0&42.5\\
$50$ &69.5&\textbf{61.8}&\textbf{44.0}\\
$100$ &69.2&61.3&41.5\\
$200$ &68.5&60.8&42.8\\
$300$ &67.4&59.7&42.7\\
		\bottomrule
	\end{tabular}
    \caption{
    Ablation on number of object queries.
    }
	\label{tbl:ablation_num_queries}
	\vspace{-0.35cm}
\end{table}

\subsection{Analysis of the Effect of TSVS}
\label{tsvs_effect_analysis}
To further illustrate and analyze the effect of the \textit{temporal segment voting scheme} (TSVS), we refer to the zebras example in the third row of \cref{fig:example} and the orange text query. Without TSVS, a prediction would have to be made for each frame in the video \textit{separately}. Hence, as the correct zebra (marked in orange) is not yet visible in the first two frames, one of the other visible zebras in each of these frames may be wrongly selected. With TSVS, however, the predictions of the correct zebra in the final three frames vote \textit{together} as part of a sequence, and due to the high reference scores of these predictions, this sequence is then selected over all other instance sequences. This results in only the correct zebra being segmented throughout the video (i.e., no zebra is segmented in the first two frames), as expected.

\end{document}